\documentclass{article}
\usepackage{spconf,amsmath,amssymb,graphicx,booktabs}
\usepackage[font=small,labelfont=bf,skip=2pt]{caption}
\usepackage{xcolor}
\usepackage{url}
\usepackage{eso-pic}
\newcommand{\abs}[1]{\lvert #1\rvert}

\title{NormMoX: A Normative Mixture of Experts for Explainable Autism Classification from Structural and Functional MRI}
\name{Muhammad Hasan Masrur$^{1,2}$, Fani Deligianni$^1$}
\address{$^{1,2}$ School of Computing Science, $^2$ James Watt School of Engineering, University of Glasgow, UK 
}
\usepackage{float}
\usepackage{stfloats}
\usepackage[hidelinks]{hyperref}
\begin{document}
\AddToShipoutPictureFG*{\AtPageUpperLeft{%
  \put(\LenToUnit{0.55in},\LenToUnit{-0.62in}){\parbox[t]{7.4in}{\centering\footnotesize
  This work has been submitted to the IEEE for possible publication. Copyright may be transferred without notice, after which this version may no longer be accessible.}}}}
\raggedbottom
\maketitle

\begin{abstract}

While structural and functional magnetic resonance imaging (sMRI, fMRI) offer profound potential for researching autism spectrum disorder (ASD) and understanding individual neurodivergence, current classifiers lack the explanatory power needed for scientific study, clinical use and personal-level insights.
Most models learn from
imaging features that also vary with age, sex, scanner site and head motion. They fuse the two modalities without a common
reference and explain decisions against arbitrary reference inputs rather than a typical brain. We present NormMoX, a normative
mixture of experts that makes the typical brain the shared reference for prediction, fusion and explanation. Normative models
trained on large healthy cohorts give the value each brain measure is expected to take for a person's age and sex. NormMoX learns
from each person's deviations from these expected values. For each modality, a neural network expert turns the deviations into a
score of evidence for ASD. A fusion head, MoX, compares each score with those of typical controls before combining them. It
therefore needs no imputation when a modality is missing and shows in which modality each person is atypical. On
1,035 participants of ABIDE~I, NormMoX reached an AUROC of $0.728\pm0.004$ in repeated cross-validation, above linear,
single-modality and deep fusion baselines. NormMoX thus shows not only whether a person is likely to have ASD, but
where, and in which modality, their brain departs from typical.
The implementation is available at \url{https://github.com/Novelex/NormMox}.
\end{abstract}


\section{Introduction}
\label{sec:intro}
Many MRI classifiers have been proposed for ASD, yet none has reached clinical use \cite{Hiremath2021TP}. However, a prediction alone gives a clinician little
to act on. The model must also highlight which brain regions are atypical, in which modality and by how much \cite{VanderVelden2022MIA}. This is challenging in ASD and other heterogeneous neurodevelopmental conditions, whose
brain differences vary from one person to the next~\cite{zabihi2019bpcni}.

By relying on raw measures, most classifiers risk capturing age, sex, scanner site and head motion effects rather than ASD~\cite{artiles2023neuroinfo}. Normative modelling avoids this \cite{Rutherford2022NP} by expressing each measure as a deviation
from what is expected for the person's age, sex and site in a healthy population. In ASD, however, norms are mostly estimated from a few hundred controls of the same study ~\cite{zabihi2019bpcni,shon2025tmi} in one
modality~\cite{pinaya2019hbm}. Lifespan models pretrained on large external healthy datasets now provide baselines for both
structural and functional MRI~\cite{rutherford2022elife,rutherford2023elife}. Multimodal models add two further challenges: their fusion does not show in which modality a person is atypical, and their
explanations measure each input's effect against an arbitrary reference input rather than a typical
brain~\cite{Pang2026EngMed,sundararajan2017icml}.

\begin{figure*}[t]
\centering
\includegraphics[width=\textwidth]{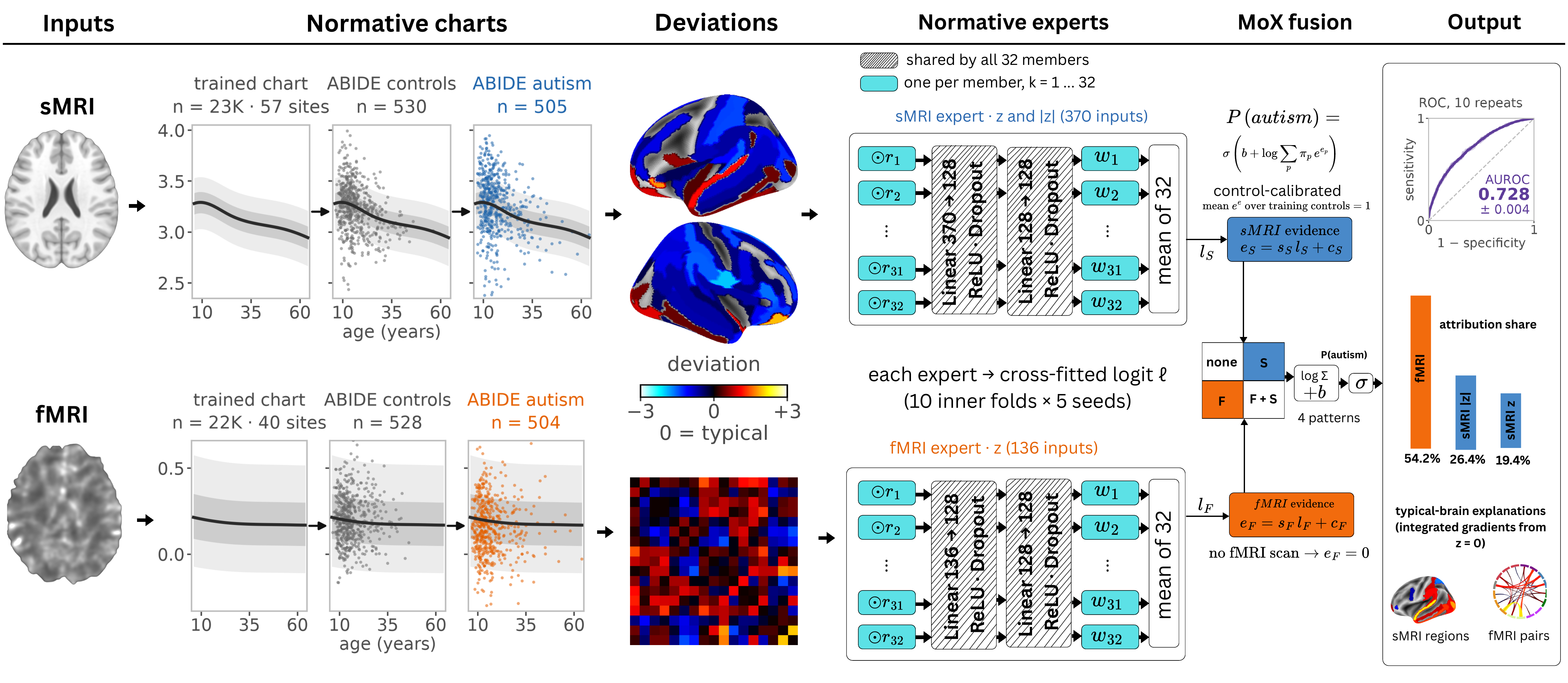}
\caption{Overview of NormMoX. Normative models turn sMRI and fMRI measures into deviations from typical brain. One
expert per modality scores the deviations, MoX fuses the scores relative to typical controls, and integrated gradients from the
typical brain explain predictions.}
\label{fig:pipeline}
\end{figure*}

We present NormMoX (Fig.~\ref{fig:pipeline}), a normative mixture of experts that classifies ASD from structural MRI (thickness and volume) and functional MRI (connectivity) and explains each decision relative to the typical brain. To our knowledge, it is the first ASD classifier built on
lifespan-model deviations of both modalities. We make four contributions: (i)~We formulate multimodal classification in a \emph{normative reference frame} where lifespan models of
healthy brains define the typical brain, providing NormMoX with a shared reference for learning, fusing and explaining data. (ii)~\emph{Normative experts} for each modality use an ensemble of multilayer perceptrons with shared weights to score how strongly deviations from age and sex norms indicate ASD. Averaging the predictions of these perceptrons keeps the score stable
on small multi-site cohorts. (iii)~We introduce \emph{MoX}, a fusion head that first measures each expert's score against typical controls and then combines
the two. Unlike learned gating~\cite{jacobs1991nc}, it makes the modalities comparable, needs no imputation for a missing modality
and shows in which modality each person is atypical. (iv)~We anchor explanations to the typical brain (\emph{typical-brain attribution}) rather than arbitrary reference inputs \cite{Sturmfels2020Distill}. Integrated gradients reveal exactly 
which regions and connections drive NormMox decision and in which modality. On 1,035 participants from 20 ABIDE sites, NormMoX reached an AUROC of
0.728, outperforming linear, single-modality and deep fusion baselines, and retained 0.711 on unseen sites. 

\section{Method}
\label{sec:method}
Fig.~\ref{fig:pipeline} outlines the NormMoX framework. Normative models first map brain measurements into deviations from expected
values (Sec.~\ref{ssec:enc}). Modality-specific  experts then evaluate these deviations as evidence for ASD
(Sec.~\ref{ssec:experts}). Finally, MoX compares these scores against typical controls and fuses both modalities (Sec.~\ref{ssec:mox}), while
integrated gradients provide explainability for each prediction (Sec.~\ref{ssec:xai}).

\subsection{Normative modelling}
\label{ssec:enc}

We quantify individual deviations from healthy lifespan norms in cortical thickness, brain volumes~\cite{rutherford2022elife} and functional connectivity~\cite{rutherford2023elife}. The normative models were fitted to external healthy reference datasets \cite{rutherford2022elife,rutherford2023elife}. Application to new acquisition sites in the Autism Brain Imaging Data Exchange (ABIDE I) dataset is achieved by calibrating site-specific mean and variance using local reference controls.

Warped Bayesian linear regression \cite{Fraza2021NI} is adopted to predict the expected value $\mu_j$ of each measurement $y_j$ (a thickness, volume or connectivity value). Measurements are first transformed to an approximately Gaussian scale via a sinh-arcsinh function  $\phi_j$. B-splines of age capture complex, non-monotonic lifespan trajectories. Additional covariates include sex, site and, for fMRI, head motion.
Residuals are computed as $u_j=\phi_j(y_j)-\mu_j$. To adjust for site-specific baseline differences, $u_j$ is centered and scaled relative to site reference controls: $z_j=(u_j-\bar{u}_j) / d_j$
where $\bar{u}_j$ and $d_j$ are the mean and standard deviation of the control residuals. The resulting $z_j$ measures site-calibrated deviation from the normative baseline ($z=0$) in standard deviation units.

\subsection{Normative experts}
\label{ssec:experts}
Each modality $m\in\{F,S\}$ (fMRI, sMRI) is processed by a dedicated ensemble of multi-layer perceptrons (MLPs), or normative expert, which maps input $z$-scores (Sec.~\ref{ssec:enc}) to a modality-specific ASD evidence score. 
Decoupling these scores enables weighted multi-modal fusion
(Sec.~\ref{ssec:mox}), while isolating their individual contributions for modality-specific attribution (Sec.~\ref{ssec:xai}). The fMRI expert receives the connectivity deviations, $x_F=z_F$. To capture bidirectional structural alterations in ASD~\cite{zabihi2019bpcni}, the sMRI expert receives both signed deviations and their absolute values, $x_S=(z_S,\abs{z_S})$. This provides an explicit inductive bias, allowing the network to evaluate deviation magnitude independently of direction.

Hidden layers enable each expert to combine deviations non-linearly. To address limited training data, each expert ensembles $K$ members, reducing sensitivity to random initialisation ~\cite{Lakshminarayanan2017NeurIPS}.
 These members share hidden layers with dropout, $g_m$, keeping the ensemble comparable in size to a single network ~\cite{gorishniy2025arxiv}. Each member $k$ adds only its own input scaling
$r_k$, output weights $w_k$ and bias $\beta_k$:
\begin{equation}
p_m=\frac{1}{K}\sum_{k=1}^{K}\sigma\big(w_k^{\top}g_m(r_k\odot x_m)+\beta_k\big),
\label{eq:expert}
\end{equation}
where $\sigma$ is the sigmoid function, $\odot$ is element-wise multiplication and $p_m$ is the members' average predicted
probability of ASD. Each member is trained with binary cross-entropy on its own prediction, so the members stay different and
their average varies less than any single member. The expert's score is $\ell_m=\log\big(p_m/(1-p_m)\big)$, positive when
$p_m>0.5$ and negative otherwise.

\subsection{Reference-calibrated mixture (MoX)}
\label{ssec:mox}
MoX fuses the two expert scores, $\ell_F$ and $\ell_S$ (Sec.~\ref{ssec:experts}), into one score $f$ in two steps.
First, because the two experts score on different scales, MoX makes them comparable by rescaling each with a learned positive factor $s_m$ and shifting it by a control reference $c_m$, computed from the training controls $\mathcal{C}$:
\begin{equation}
e_m=s_m\ell_m+c_m,\qquad c_m=-\log\frac{1}{|\mathcal{C}|}\sum_{i\in\mathcal{C}}e^{s_m\ell_{m,i}}.
\label{eq:evidence}
\end{equation}
Since $e^{e_m}$ averages to one over controls, $e_m$ acts as log-scale evidence for ASD against the controls.
The evidence $e_m$ is zero at the control reference and positive when the participant scores above it, towards ASD.

Second, a participant may show evidence in fMRI, in sMRI, in both or in neither, and MoX gives each case its own term:
\begin{equation}
f=b+\log\big(\pi_0+\pi_F e^{e_F}+\pi_S e^{e_S}+\pi_B e^{e_F+e_S}\big),
\label{eq:mox}
\end{equation}
where $b$ is a bias and the learned weights $\pi_0$, $\pi_F$, $\pi_S$ and $\pi_B$ are positive, sum to 1 and correspond to no evidence, fMRI only, sMRI only and both. Because each modality has its own term, strong evidence in one modality raises $f$ even when the other shows little. The prediction is $\sigma(f)$. A participant without fMRI receives $e_F=0$ and is classified from the sMRI evidence alone.

\subsection{Attribution from the typical brain}
\label{ssec:xai}

\begin{figure*}[t]
\centering
\begin{minipage}[t]{0.49\textwidth}
\centering
\includegraphics[width=\linewidth,height=12in,keepaspectratio]{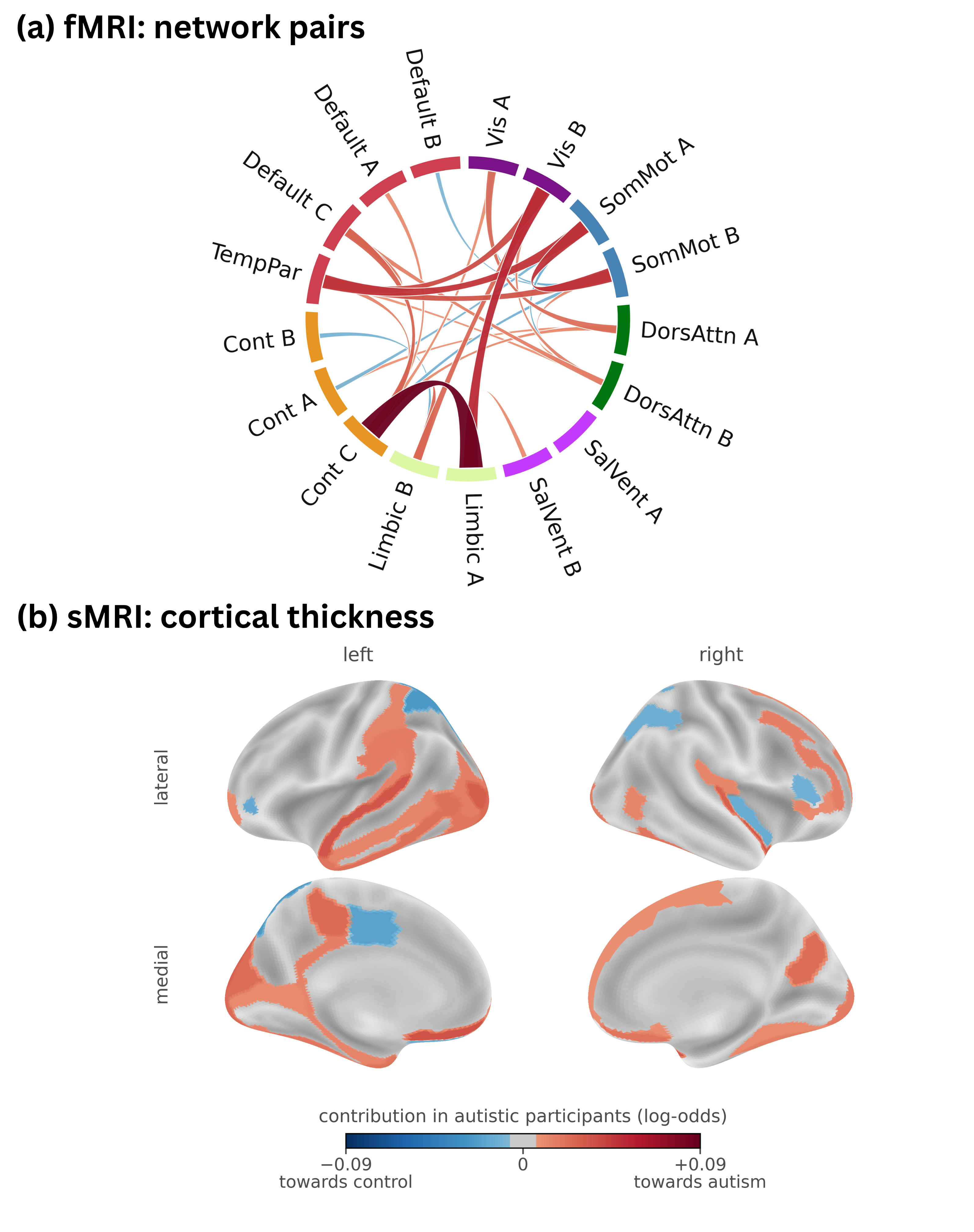}
\caption{Mean contribution to the fused prediction in autistic test participants, where consistent across folds ($\lvert$mean$\rvert\ge2$ SD). Red: towards ASD; blue: towards control. (a) Ribbon width: contribution of each network pair. (b) sMRI: cortical thickness on inflated surfaces.}
\label{fig:maps}
\end{minipage}\hfill
\begin{minipage}[t]{0.49\textwidth}
\centering
\includegraphics[width=\linewidth,height=4.7in,keepaspectratio]{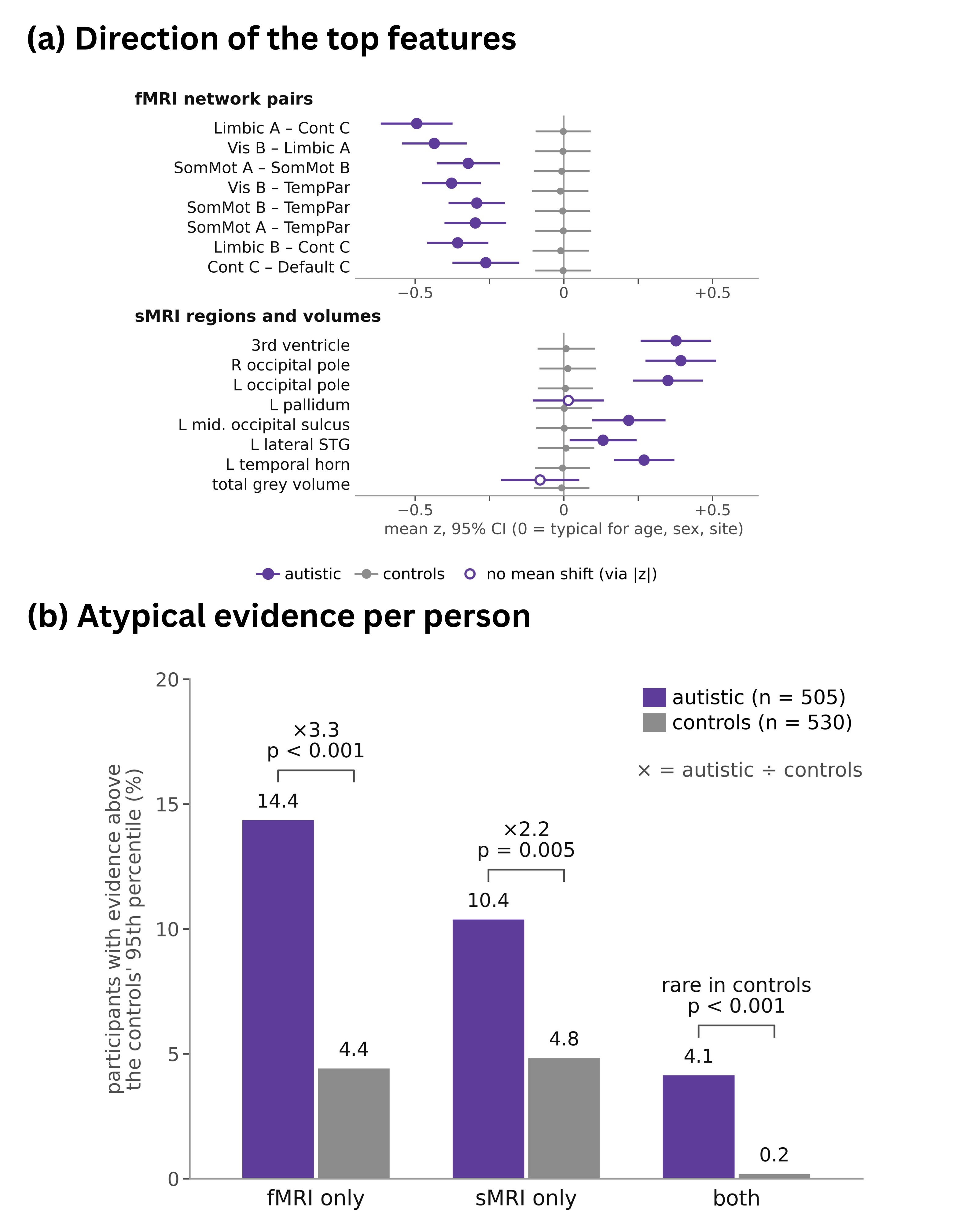}
\caption{(a) Mean $z$ (95\% CI) of the most separating features; open circles: no significant mean shift (contribution through $\abs{z}$). (b) Participants flagged in fMRI only, sMRI only or both (10-repeat mean).}
\label{fig:feat}
\end{minipage}
\end{figure*}

To explain each prediction, we measure how much each deviation
moves the MoX score $f$ away from its value for the typical brain, $x^0=0$. We use integrated gradients~\cite{sundararajan2017icml}, which gradually change the input from $x^0$ to
the participant's deviations $x$ and accumulate the gradient of $f$ along it:
\begin{equation}
A_j(x)=x_j\int_0^1\frac{\partial f(\alpha x)}{\partial x_j}\,d\alpha .
\label{eq:ig}
\end{equation}
A positive $A_j$ moves the prediction towards ASD, and a negative $A_j$ towards control. By completeness, the attributions sum to  $f(x)-f(0)$, the participant's total departure from the typical brain. Averaged over ASD participants, they map the deviations that drive the predictions (Fig.~\ref{fig:maps}). 

\section{Experiments}
\label{sec:exp}
We used all 1,035 ABIDE~I participants (505 with ASD,
530 controls, 20 site samples) \cite{dimartino2014mp}. Each participant has 185 FreeSurfer~5.1  structural measures (Destrieux
thicknesses~\cite{destrieux2010ni} and volumes) and 136 functional correlations between the 17 Yeo networks~\cite{yeo2011jn}
(C-PAC, no band-pass filtering or global signal regression). Each expert has $K=32$ members sharing two hidden layers of 128 ReLU units (dropout 0.1). Performance was estimated with ten repeats of 10-fold
cross-validation, stratified by diagnosis and site, and every step was fitted on training data only. Each site was calibrated by
its training controls, cross-fitted so that no control was its own reference, and MoX was fitted on out-of-fold expert scores
(10 inner folds, five seeds). Leave-one-site-out evaluation, with each held-out site calibrated by its own controls, tests
transfer to new sites. NormMoX rests on three design choices: normative deviations as inputs, fusion of two modality experts, and a fusion head on trained experts. We assess each against baselines on the same splits. 

We ablated the representation of the brain measurements by training logistic
regression on the same structural features and functional connectivity in three ways: as raw values, as values normed within
ABIDE by each site's training controls (ABIDE-normed), or as normative deviations. We also ablated the modality by evaluating each expert alone. We compared NormMoX with two deep fusion baselines that replace the sMRI expert with a transformer under the MoX head. In the first, the experts are trained separately and frozen while only the head is fitted,  which assesses the transformer as an
expert. In the second, experts and head are trained together end to end (Table~\ref{tab:perf}). 

Finally, we evaluated the explanations on four criteria (Table~\ref{tab:xai}). Faithfulness is measured by the drop in AUROC when inputs are reset to the typical brain ($z=0$), stability by the agreement of attribution maps across folds, specificity by the attribution left when labels are shuffled within site \cite{Nauta2023ACMCS}, and calibration by the flag rate in controls, which should match its $5\%$ target. AUROC differences were tested with the corrected resampled $t$-test of Nadeau and Bengio~\cite{nadeau1999nips}, which accounts for overlapping training sets, with Holm correction. 

\section{Results}
\label{sec:results}

\noindent\textbf{Classification.} NormMoX achieved the best AUROC ($0.728\pm0.004$), AUPRC, balanced accuracy, specificity and F1
(Table~\ref{tab:perf}).
The ablation experiments demonstrated that normative deviations outperformed raw and ABIDE-normed inputs, while fusion bested single modalities. NormMoX also outperformed both deep fusion baselines. Its gains over the single experts and over raw and ABIDE-normed inputs were significant after Holm correction. On unseen sites, NormMoX kept an AUROC of 0.711, above both experts (fMRI 0.690, sMRI 0.633).

\begin{table}[H]
\centering
\caption{Classification on ABIDE~I over 10 repeats of 10-fold cross-validation (AUROC $\pm$ SD over repeats).}
\label{tab:perf}
\footnotesize
\setlength{\tabcolsep}{1.9pt}
\begin{tabular}{@{}lcccccc@{}}
\toprule
Model & AUROC & AUPRC & BAcc & Sens & Spec & F1 \\
\midrule
\multicolumn{7}{@{}l}{\textit{Input ablation: logistic regression}} \\
Raw features & $0.682\pm0.005$ & 0.675 & 0.632 & 0.607 & 0.657 & 0.616 \\
ABIDE-normed & $0.685\pm0.007$ & 0.678 & 0.634 & 0.614 & 0.654 & 0.620 \\
Normative deviations & $0.691\pm0.008$ & 0.688 & 0.637 & \textbf{0.619} & 0.654 & 0.624 \\
\multicolumn{7}{@{}l}{\textit{Modality ablation: single expert}} \\
sMRI ($z$, $\abs{z}$) & $0.652\pm0.005$ & 0.651 & 0.602 & 0.517 & 0.686 & 0.559 \\
fMRI ($z$) & $0.699\pm0.005$ & 0.697 & 0.643 & 0.605 & 0.681 & 0.623 \\
\multicolumn{7}{@{}l}{\textit{Deep fusion baselines}} \\
Experts frozen & $0.708\pm0.007$ & 0.714 & 0.644 & 0.583 & 0.704 & 0.615 \\
Jointly fine-tuned & $0.711\pm0.007$ & 0.717 & 0.645 & 0.586 & 0.703 & 0.617 \\
\midrule
\textbf{NormMoX} & $\mathbf{0.728\pm0.004}$ & \textbf{0.730} & \textbf{0.658} & 0.604 & \textbf{0.712} & \textbf{0.633} \\
\bottomrule
\end{tabular}
\end{table}

\noindent\textbf{Interpretability.} NormMoX's explanations were faithful, stable, specific and
calibrated (Table~\ref{tab:xai}). Functional evidence was focal (Fig.~\ref{fig:maps}a). The leading network pairs, headed by Limbic~A--Control~C, were all hypoconnected in ASD (Fig.~\ref{fig:feat}a), in line with earlier ABIDE findings~\cite{dimartino2014mp}. Structural evidence was distributed (Fig.~\ref{fig:maps}b), with enlarged ventricles and thicker occipital cortex \cite{Bedford2025BP} but mostly deviations in both directions that cancel out in group averages \cite{Fan2025CB}. Most flagged participants with ASD had evidence in only one modality, so either modality alone would miss a third to a half of them (Fig.~\ref{fig:feat}b).

\begin{table}[H]
\centering
\caption{Explanation checks (100 folds); $^{*}p<0.05$, $^{***}p<0.001$.}
\label{tab:xai}
\footnotesize
\setlength{\tabcolsep}{2.5pt}
\begin{tabular}{@{}lccc@{}}
\toprule
 & fMRI (136) & sMRI (370) & All (506) \\
\midrule
\multicolumn{4}{@{}l}{\textit{Faithful:} AUROC (full model 0.728)} \\
All inputs reset to typical & $0.651^{***}$ & $0.699^{***}$ & $0.499^{***}$ \\
Extra drop, top-30 vs random-30 & $0.063^{***}$ & $0.014^{*}$ & $0.064^{***}$ \\
\multicolumn{4}{@{}l}{\textit{Stable across folds}} \\
Spearman $\rho$ of maps & 0.71 & 0.76 & 0.75 \\
Top-10 overlap / same sign & 0.79 / 1.00 & 0.69 / 1.00 & 0.71 / 1.00 \\
\multicolumn{4}{@{}l}{\textit{Specific:} shuffled labels (2 folds)} \\
Attribution left & 6--7\% & $<$0.02\% & 3--4\% \\
\multicolumn{4}{@{}l}{\textit{Evidence flags} (All: either modality)} \\
Controls flagged (target) & 4.6\% (5\%) & 5.0\% (5\%) & 9.4\% (9.75\%) \\
Autistic flagged (Fisher test) & $18.5\%^{***}$ & $14.5\%^{***}$ & $28.9\%^{***}$ \\
\bottomrule
\end{tabular}
\end{table}

\section{Conclusion}
\label{sec:conclusion}
We presented NormMoX, which uses the typical brain as one common reference for classifying ASD from structural and functional MRI, fusing the two modalities and explaining each decision. NormMoX outperformed linear, single-modality and deep fusion baselines, transferred to unseen sites and showed where, and in which modality, each person departs from typical. Because the normative models stay fixed, a new site needs only its own controls. NormMoX can also be extended to other neurodevelopmental disorders.
\raggedbottom
\section{Compliance with ethical standards}
This study was conducted retrospectively using open-access human data from ABIDE~\cite{dimartino2014mp}. Ethical approval was not required, as confirmed by the licence attached to the data.

\section{Acknowledgments}
This work was funded by the Engineering and Physical Sciences Research Council (EP/W524359/1). The authors declare no conflicts of interest.

\small
\bibliographystyle{IEEEbib}
\bibliography{refs}
\end{document}